\documentclass[conference,a4paper]{IEEEtran}
\IEEEoverridecommandlockouts

\usepackage[hidelinks]{hyperref}
\usepackage[cmex10]{amsmath}%American Math Society(AMS) math formatting
\usepackage{amssymb,amsfonts}%AMS extra symbols and fonts
\usepackage{dblfloatfix}%fix double column figure ordering and placement

\usepackage[ruled,vlined]{algorithm2e}
\usepackage{graphicx}
\graphicspath{{Figures/PDF/}{Figures/PNG/}}

\usepackage{booktabs}
\usepackage{siunitx}
\usepackage[numbers,compress]{natbib}
\usepackage{texnames}
\usepackage{bm,bbm}
\usepackage{orcidlink}

\begin{document}

\title{\uppercase{Comparative Assessment of multimodal earth observation data for Soil Moisture Estimation}
% \thanks{This research is funded by the European Union’s Horizon Europe Programme under Grant Agreement No. 101135422 (UNIVERSWATER)}
}

% \author{	
%     \IEEEauthorblockN{Alejandro C.\ Frery\orcidlink{0000-0002-8002-5341}}
% 	\IEEEauthorblockA{\textit{Victoria University of Wellington}\\
% 		6140 Wellington, New Zealand\\
% 		alejandro.frery@vuw.ac.nz}
        
% 	\and
% 	\IEEEauthorblockN{Hui Zhang\orcidlink{0000-0002-5283-7350}}
% 	\IEEEauthorblockA{\textit{Inner Mongolia University}\\
% 		010021 Hohhot, China\\
% 		hui.zhang@imu.edu.cn}
% 	\and
% 	\IEEEauthorblockN{Andrea Rey\orcidlink{0000-0002-9185-1382}}
% 	\IEEEauthorblockA{\textit{Universidad Nacional de Hurlingham}\\
% 		1688 República Argentina\\
% 		andrea.rey@unahur.edu.ar}
% }

\author{
    Ioannis Kontogiorgakis\textsuperscript{1},
    Athanasios Askitopoulos\textsuperscript{1},
    Iason Tsardanidis*\textsuperscript{1},
    \thanks{*Correspondence to  \href{mailto:j.tsardanidis@noa.gr}{j.tsardanidis@noa.gr}}
    Dimitrios Bormpoudakis\textsuperscript{1},\\
    Ilias Tsoumas\textsuperscript{1,2},
    Fotios Balampanis\textsuperscript{1},
    Charalampos Kontoes\textsuperscript{1}\\[1em]
    
    \textsuperscript{1}BEYOND EO Centre, IAASARS, National Observatory of Athens, Athens, Greece\\
    \textsuperscript{2}Artificial Intelligence, Wageningen University \& Research, The Netherlands\\
}

% TODO:
% - Add SM mean in order to explain MAE -> resolved
% - Fix Ex2: 16 or 10 window -> resolved

% Other titles
% EVALUATING FOUNDATION MODEL EMBEDDINGS VERSUS TRADITIONAL FEATURES FOR HIGH-RESOLUTION SOIL MOISTURE ESTIMATION ACROSS EUROPE

% FOUNDATION MODELS VS. HAND-CRAFTED FEATURES FOR MULTIMODAL SOIL MOISTURE ESTIMATION

% MULTIMODAL SOIL MOISTURE ESTIMATION: EVALUATING PRITHVI EMBEDDINGS AND TEMPORAL FUSION STRATEGIES

\maketitle
\begin{abstract}
	Accurate soil moisture (SM) estimation is critical for precision agriculture, water resources management and climate monitoring. Yet, existing satellite SM products are too coarse ($>$1 km) for farm-level applications. We present a high-resolution (10 m) SM estimation framework for vegetated areas across Europe, combining Sentinel-1 SAR, Sentinel-2 optical imagery and ERA-5 reanalysis data through machine learning. Using 113 International Soil Moisture Network (ISMN) stations spanning diverse vegetated areas, we compare modality combinations with temporal parameterizations, using spatial cross-validation, to ensure geographic generalization. We also evaluate whether foundation model embeddings from IBM-NASA's Prithvi model improve upon traditional hand-crafted spectral features. Results demonstrate that hybrid temporal matching—Sentinel-2 current-day acquisitions with Sentinel-1 descending orbit—achieves R²=0.514, with 10-day ERA5 lookback window improving performance to R²=0.518. Foundation model (Prithvi) embeddings provide negligible improvement over hand-crafted features (R²=0.515 vs. 0.514), indicating traditional feature engineering remains highly competitive for sparse-data regression tasks. Our findings suggest that domain-specific spectral indices combined with tree-based ensemble methods offer a practical and computationally efficient solution for operational pan-European field-scale soil moisture monitoring.

    % changed 'crop lands, tree cover, grasslands, sparse vegetated areas' -> 'diverse vegetated' in abstract in order to properly mention them in Ground Truth Data section 
    
    % Results demonstrate that combining Prithvi-derived optical features with Sentinel-1 backscatter and 7-day ERA5 meteorological history, significantly outperform raw-satellite bands (R²=0.49 vs. 0.46, +6.5\% improvement \textbf{(TODO change this)}. Our hybrid framework—frozen foundation model encoder with classical machine learning—balances accuracy and computational efficiency, providing a scalable pathway for pan-European field-scale soil moisture monitoring systems.
\end{abstract}

\begin{IEEEkeywords}
	Soil Moisture Estimation, ISMN, Machine Learning, Foundation Model.
\end{IEEEkeywords}

\section{Introduction}

Soil moisture (SM) is a critical variable in the Earth system, directly influencing agricultural productivity, hydrological processes, and climate dynamics \cite{PENG2021112162}. Accurate, high-resolution SM estimates are essential for precision agriculture, drought monitoring, and irrigation management \cite{PILES2016403, essd-11-1583-2019}. While satellite-based SM products exist at coarse spatial resolutions (e.g., SMAP at 9-36 km \cite{CHAN2018931}, SMOS at 25-50 km \cite{5446359}), agricultural applications increasingly demand finer spatial detail to capture field-scale heterogeneity \cite{Montzka2020SoilMP, sadri2020global}.

Ground-based measurements from networks such as the International Soil Moisture Network (ISMN) provide accurate point observations but suffer from sparse spatial coverage \cite{hess-25-5749-2021, hess-15-1675-2011}. Traditional SM retrieval methods rely on physics-based models or empirical relationships \cite{ELHAJJ2016202, article}, yet struggle to integrate multiple data sources effectively and capture complex, nonlinear relationships between surface conditions and soil water content \cite{rs70810098}. Machine learning (ML) approaches have demonstrated the potential for SM estimation by learning patterns from multimodal satellite observations, with tree-based methods (Random Forest, XGBoost) proven effective in integrating optical, SAR and meteorological data \cite{article2,article3, VULOVA2021147293}, and deep learning capturing spatial-temporal dependencies \cite{NearRealTimeForecastofSatelliteBasedSoilMoistureUsingLongShortTermMemorywithanAdaptiveDataIntegrationKernel,fang2017prolongation, article6} 

% [23]–[25].

\begin{figure}[hbt]
	\centering
	\includegraphics[width=1\linewidth]{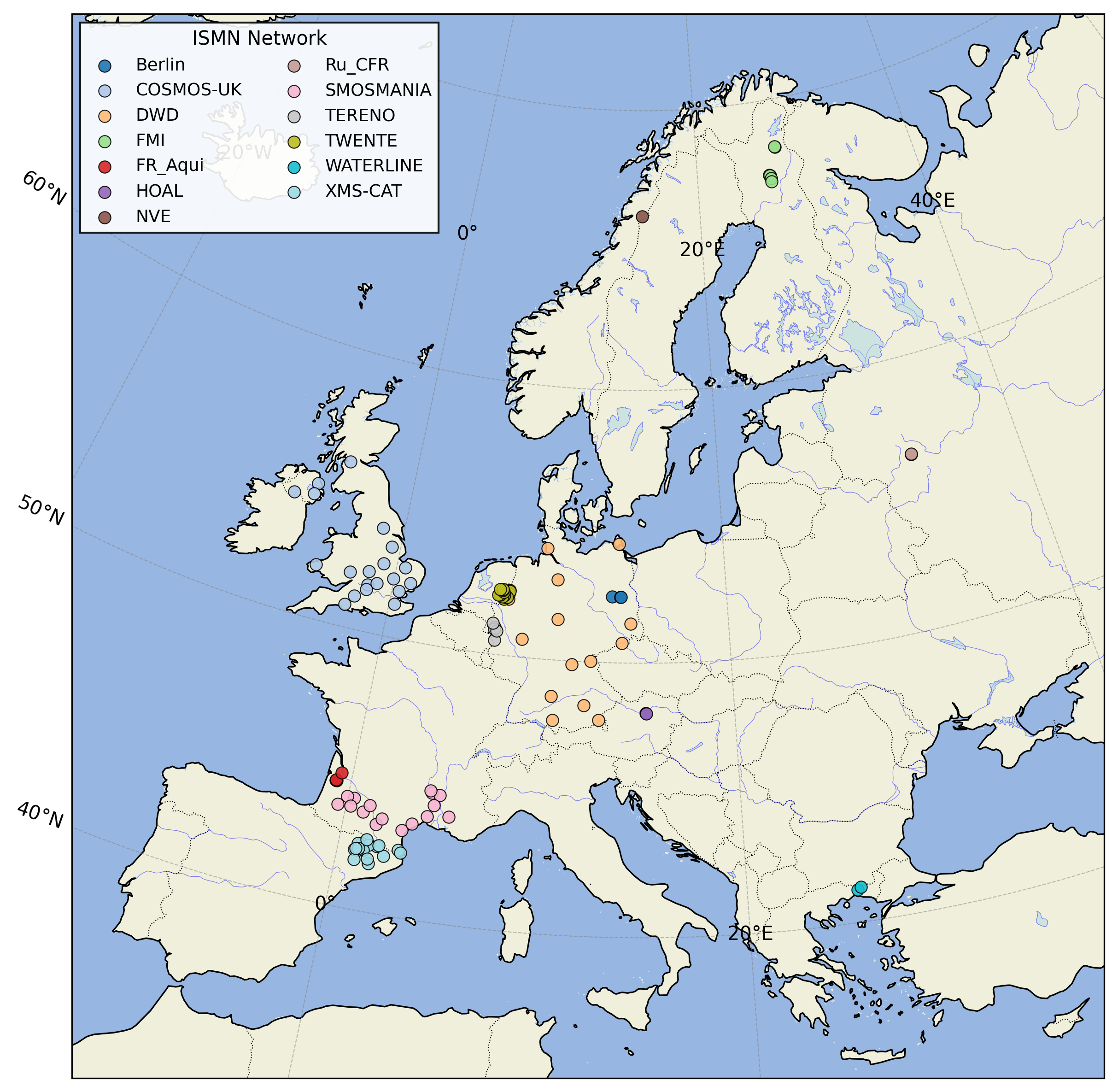}
	\caption{Distribution of ISMN stations across Europe used in this study after filtering, colored by network.}\label{fig:stations_location}
\end{figure}

Recent geospatial foundation models, such as Prithvi \cite{unknown}, offer pre-trained feature extractors learned from vast amounts of satellite imagery using masked autoencoder architectures on Harmonized Landsat Sentinel-2 (HLS) data \cite{tseng2023lightweight,cong2022satmae}. While these models have demonstrated success in segmentation and classification \cite{article6,yuan2020self}, their effectiveness for regression tasks with limited ground truth—particularly compared to classical ML with engineered features—remains unexplored for SM applications. This gap is critical as operational systems require robust performance under data-scarce conditions typical of in-situ monitoring networks.

In this study, we develop and evaluate a multimodal approach for high-resolution SM estimation across Europe using 113 ISMN stations (2019-2025), combining Sentinel-2 optical imagery, Sentinel-1 SAR data and ERA5 meteorological reanalysis. We systematically compare traditional ML methods with hand-crafted features against approaches utilizing pre-trained Prithvi embeddings. Our contributions are threefold: (1) we demonstrate that hybrid temporal matching strategies—Sentinel-2 current-day acquisitions with Sentinel-1 descending orbit—achieve optimal performance, with descending SAR consistently outperforming ascending configurations; (2) we identify that 10-day ERA5 lookback window outperforms both shorter and longer integration periods across the 0-20 day range and (3) we show that foundation model embeddings provide negligible improvement over hand-crafted features (R²=0.515 vs 0.514), indicating traditional feature engineering remains highly competitive for sparse-data regression tasks. Our findings suggest that domain-specific spectral indices combined with optimal temporal parameterization offer a practical and computationally efficient pathway for operational deployment, even with limited training data. 
% Our contributions are threefold: (1) we demonstrate that Prithvi embeddings outperform hand-engineered satellite features, achieving 6.5\% improvement (R²=0.49 vs 0.46); (2) we quantify incremental contributions of each modality, showing ERA5 provides the largest gain while SAR and topography offer consistent improvements; and (3) we provide practical guidance for operational systems by identifying optimal temporal parameterization (7-day ERA5 window) and feature fusion strategies. Our findings indicate that hybrid frameworks that combine the optical representations derived from the foundation model with traditional meteorological features achieve the best performance, offering a practical deployment pathway even with limited training data.

\section{Data Collection}
    \subsection{Ground Truth Data}
    Data were collected from the International Soil Moisture Network (ISMN) [9], [10], a global repository of in-situ soil moisture measurements from operational monitoring networks. We extracted daily soil moisture observations at 5 cm depth from 113 stations across Europe, spanning the period from January 2019 to December 2024. Each measurement record contains a station identifier, timestamp, volumetric soil moisture value (m³/m³), and geographic coordinates. Stations were filtered to include only vegetated land cover types - crop lands, tree cover, grasslands and sparse vegetated areas. To ensure station independence, we removed stations within 1km of each other, reducing the dataset to 113 spatially independent locations. Table \ref{tab:ISMN_stations} reports the ISMN networks used in this study, including station counts, number of measurements, and mean soil moisture for each network.

\begin{table}[h]
	\centering
	\caption{Statistics of ISMN networks used in this study.}
	\label{tab:ISMN_stations}
	\begin{tabular}{@{}cccc@{}}  % @{} removes left/right padding, r for right-aligned numbers
		\toprule
		\textbf{Network} & \textbf{\# Stations} & \textbf{\# Meas.} & \textbf{Mean SM}\\
		\midrule
		Cosmos-UK   & 20 & 1399 & 0.29\\
		TWENTE      & 18 &  773 & 0.20\\
		XMS-CAT     & 18 & 2862 & 0.14\\
		SMOSMANIA   & 16 & 2539 & 0.17\\
		DWD         & 14 &  410 & 0.22\\
		FMI         &  8 &  478 & 0.20\\ 
		TERENO      &  5 &  457 & 0.22\\
		FR\_Acqui   &  4 &  333 & 0.13\\
		Berlin      &  3 &  311 & 0.06\\
		HOAL        &  2 &   73 & 0.28\\
		Ru\_CFR     &  2 &   49 & 0.38\\
		WATERLINE   &  2 &  137 & 0.11\\
		NVE         &  1 &   10 & 0.14\\
		\bottomrule
	\end{tabular}
\end{table}

% \begin{table}[h]
% 	\centering
% 	\caption{Statistics of ISMN networks used in this study.}
% 	\label{tab:ISMN_stations}
% 	\begin{tabular}{1\columnwidth}{@{}llrr@{}}
% 		\toprule
% 		\textbf{Network} & \textbf{\# of stations} & \textbf{\# of measurements} & \textbf{mean SM}\\
% 		\midrule
% 		Cosmos-UK & 20 & 1399 & 0.2895\\
%         TWENTE & 18 & 773 & 0.1952 \\
%         XMS-CAT & 18 & 2862 & 0.1389\\
%         SMOSMANIA & 16 & 2539 & 0.1697\\
% 		DWD & 14 & 410 & 0.2198 \\
% 		FMI & 8 & 478 & 0.1953 \\ 
%         TERENO & 5 & 457 & 0.2181\\
%         FR\_Acqui & 4 & 333 & 0.1250\\
%         Berlin & 3 & 311 & 0.0596\\
%         HOAL & 2 & 73 & 0.2768\\
%         Ru\_CFR & 2 & 49 & 0.3827\\
%         WATERLINE & 2 & 137 & 0.1101 \\
%         NVE & 1 & 10 & 0.1405\\
%         % \midrule
%         % \textbf{Total Stations} & \textbf{113} \\
%         \bottomrule
% 	\end{tabular}
% \end{table}

    \subsection{Satellite Data}
    For the purpose of our analysis, we used: Sentinel-2 (S2) for optical multispectral imagery and Sentinel-1 (S1) for Synthetic Aperture Radar (SAR) observations.

    Sentinel-2 is operated by the European Space Agency (ESA) as part of the Copernicus program, providing multispectral imagery with spatial resolution up to 10 meters and temporal resolution of 5 days. We utilized S2 Level-2A (L2A) Surface Reflectance products, which are atmospherically corrected to minimize variability due to atmospheric scattering and absorption. For our analysis, we utilized all available S2 bands.

    Sentinel-1 is a C-band SAR mission providing all-weather, day-and-night imaging capability independent of cloud cover. S1 operates in Interferometric Wide (IW) swath mode with VV and VH polarizations at 10-meter spatial resolution, at a 6-12 day revisit frequency over Europe. For this study, We extracted the S1 Ground Range Detected (GRD) product. A critical consideration for S1 data is orbit geometry: ascending and descending passes observe the surface from different viewing angles, which affects backscatter sensitivity to surface roughness and moisture content. We therefore extracted S1 data in three configurations: ascending orbit only (S1\_ASC), descending orbit only (S1\_DESC), and combined ascending and descending observations (S1\_BOTH), to evaluate the impact of orbit geometry on soil moisture prediction performance.

    \subsection{Reanalysis}
    We utilized ERA5 meteorological reanalysis from the European Centre for Medium-Range Weather Forecasts (ECMWF) \cite{article7}, \cite{essd-13-4349-2021}. ERA5 provides hourly estimates of atmospheric, land, and ocean variables at approximately 25km spatial resolution, generated through data assimilation of observations into a numerical weather model. We extracted multiple daily-aggregated variables relevant to soil moisture dynamics, including:
    
    \begin{itemize}
        \item \textbf{Precipitation:} total precipitation
        \item \textbf{Temperature variables:} air, skin and soil temperature
        \item \textbf{Evapotranspiration variables:} potential evaporation 
        \item \textbf{Soil variables:} volumetric soil water content at 0-7cm
        \item \textbf{Surface variables:} surface pressure, dewpoint temperature, leaf area index
        \item \textbf{Radiation:} surface solar radiation downward, surface thermal radiation downward
        \item \textbf{Wind:} 10-meter u and v wind components
    \end{itemize}

\section{Methods}
    \subsection{Preprocessing \& Feature Engineering}

    Satellite data were temporally matched to ISMN measurements using two strategies: current day (same-date acquisitions) and closest (nearest cloud-free image within ±10 days for Sentinel-2 and Sentinel-1). Sentinel-1 was extracted in three orbit configurations (ascending, descending, combined) to evaluate viewing geometry effects. Spatial extraction used 256×256 pixel patches at 10m resolution centered on station coordinates. Sentinel-2 images with $>20\%$ cloud coverage were excluded using the Scene Classification Layer (SCL) band of S2 rasters. Meteorological variables were extracted with lookback window of 20 days from the ERA5-Land collection.

    % From Sentinel-2 collection, we calculated the indices: 
    % \begin{itemize}
    %     \item NDVI = $\frac{B_8 - B_4}{B_8 + B_4}$ 
    %     \item NDWI = $\frac{B_{8A} - B_{11}}{B_{8A} + B_{11}}$
    %     \item NDMI = $\frac{B_8-B_{11}}{B_8+B_{11}}$
    %     \item MSI = $\frac{B_{11}}{B_{8A}}$
    % \end{itemize}

    From the Sentinel-2 collection, we calculated the following vegetation and moisture indices:
    \begin{align}
    \mathrm{NDVI} &= \frac{B_8 - B_4}{B_8 + B_4} \label{eq:ndvi} \\
    \mathrm{NDWI} &= \frac{B_3 - B_8}{B_3 + B_8} \label{eq:ndwi}\\
    \mathrm{NDMI} &= \frac{B_8 - B_{11}}{B_8 + B_{11}} \label{eq:ndmi}\\
    \mathrm{MSI}  &= \frac{B_{11}}{B_{8A}} \label{eq:msi}
    \end{align}

    Sentinel-1 features included VV and VH backscatter and cross-polarization ratio (VH/VV). Temporal dynamics were captured through first-order differences and rates of change for all spectral features.   
    
    % For foundation model experiments, we firstly extracted for each available Sentinel-2 image, the 224x224 patch, centered to the ISMN station. We then normalized the patches. We extracted 768-dimensional embeddings from Prithvi's frozen encoder using the patch we extracted (500m x 500m), replacing hand-crafted Sentinel-2 bands.

    For the foundation model experiments, we extracted a 224×224 pixel Sentinel-2 image patch centered on each ISMN station for every available acquisition. The extracted patches, corresponding to an approximate ground footprint of 500m x 500m, were radiometrically normalized prior to inference. We then obtained 768-dimensional feature embeddings from the frozen encoder of Prithvi 2.0 
    foundation model (300M parameters), which were used as inputs to the downstream regression model in place of the Sentinel-2 spectral bands.
    
    \subsection{Model Training and Experiments}
    % We trained Random Forest models with default hyperparameters. Models were evaluated with a Group-5-Fold cross-validation. The dataset was split spatially at the station level, all samples from a given station are assigned either to the training set or to the test set, but never to both. This cross-validation method ensures that the model can generalize to unseen stations. Model's performance was evaluated using R², RMSE, and MAE.

    We trained Random Forest models using default hyperparameter settings. Model performance was assessed using a group-based 5-fold cross-validation strategy, in which data were partitioned at the station level, preventing any spatial leakage between sets and evaluating the model’s ability to generalize to previously unseen stations. Predictive performance was quantified using the R², RMSE and MAE.

\begin{figure}[hbt]
	       \centering
	       \includegraphics[width=0.87\linewidth]{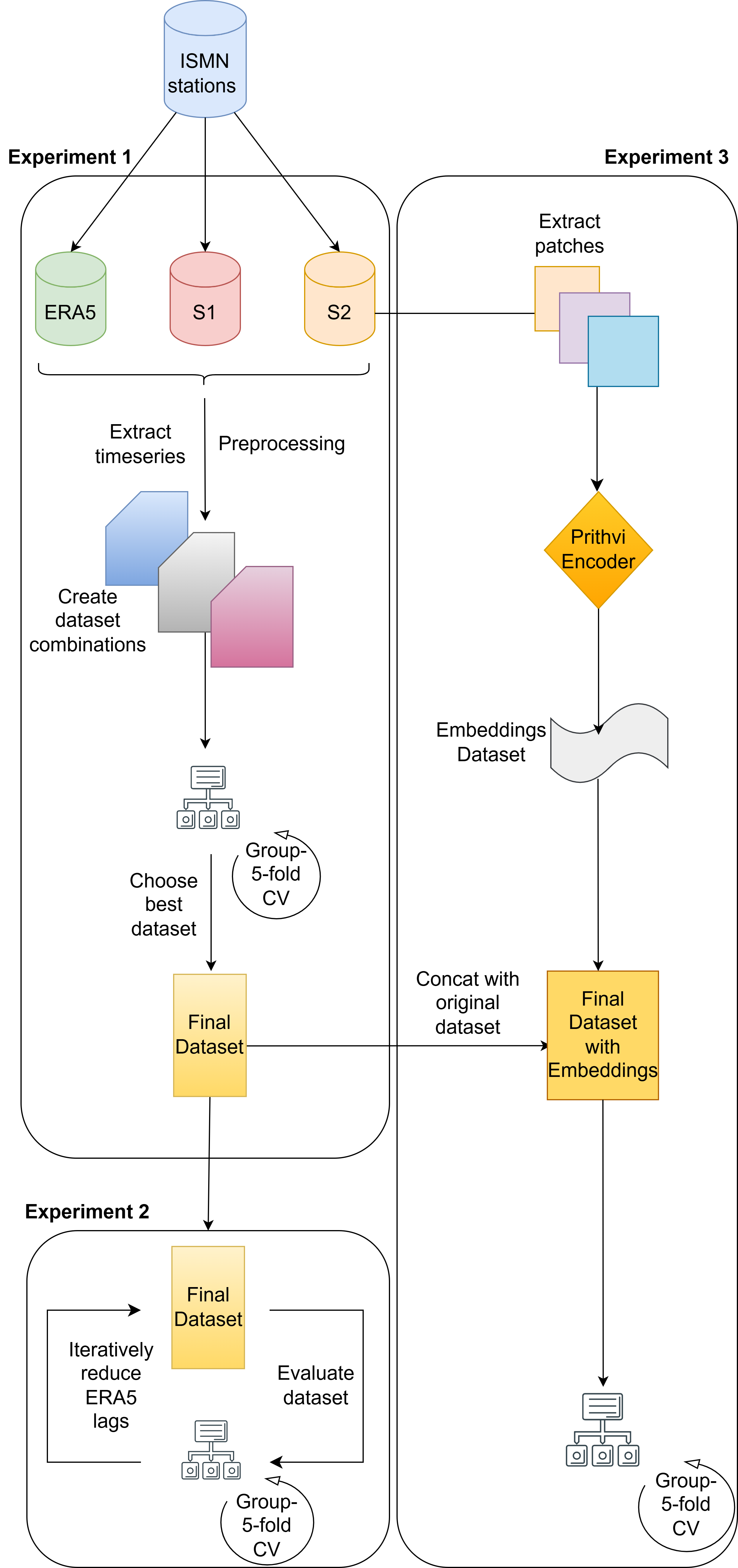}
	       \caption{Experiments methodology diagram.}\label{fig:experiments}
        \end{figure}

    Three experiments (Fig. \ref{fig:experiments}) systematically evaluated soil moisture estimation: \textbf{E1: Modality and Temporal Matching} tested individual and combined data sources (S2, S1, ERA5) with different temporal matching strategies (current day vs. closest) and S1 orbit configurations to identify optimal combinations. 
    % As shown in Fig. \ref{fig:experiments}, we evaluate all the datasets and select the best performing combination in terms of Soil Moisture estimation. 
    \textbf{E2: ERA5 Temporal Window} evaluated lookback periods (0 to 20 days) using the best configuration from E1 to optimize meteorological integration. \textbf{E3: Feature Representations} compared traditional hand-engineered features against Prithvi foundation model embeddings, testing: (a) Prithvi + ERA5 only, (b) Prithvi embeddings + S1 + ERA5 and (c) Prithvi embeddings + hand-crafted S2 indices + S1 + ERA5.

\section{Results \& Discussion} 
    \subsection{E1: Modality and Temporal Matching Evaluation}

Table \ref{tab:Experiments_scores} presents performance across sensor combinations and temporal matching strategies. All configurations include ERA5 variables (omitted from table datasets for brevity). The optimal configuration—S2 current-day with S1 DESC closest—achieved R²=0.514, RMSE=0.078 m³/m³. This hybrid temporal strategy outperformed uniform approaches: S2 closest + S1 DESC closest yielded R²=0.487, representing a 5.6\% performance loss. Current-day S2 matching preserves current surface conditions critical for soil moisture estimation, while S1's closest matching within ±10 days provides sufficient SAR context given the sensor's 6-12 day revisit cycle.

S1 orbit geometry significantly impacts performance. Descending orbit consistently outperformed ascending across all combinations (e.g., S1 DESC curr\_day: R²=0.434 vs S1 ASC curr\_day: R²=0.404), likely due to morning acquisition timing when surface moisture gradients are pronounced. Combined orbit configurations (BOTH) did not improve upon single-orbit performance, suggesting redundancy rather than complementarity between viewing geometries.

S2 alone achieved competitive performance (R²=0.507), while S1 alone performed poorly (R²=0.404-0.434), confirming optical vegetation and moisture indices are primary predictors of surface soil moisture. However, multimodal fusion consistently improved results: S2+S1 combinations gained 1-2\% R² over S2 alone, indicating SAR backscatter provides complementary information on surface roughness and moisture that optical indices cannot fully capture.

    \subsection{E2: ERA5 temporal window optimization}

Using the optimal S2 curr\_day + S1\_DESC closest configuration, we tested each ERA lookback window by continuously reducing the lag of the ERA variables. ERA5 lookback window of 10 days yielded R²,MAE and RMSE of 0.5185, 0.059 and 0.079 respectively. A negligible improvement in R² was also achieved by the 16-day ERA5 lookback window. 
This extended optimal window suggests deeper soil layers and longer-term atmospheric variables significantly influence 5cm depth measurements across European climate gradients.
% particularly in regions with low evaporative demand where precipitation effects persist beyond two weeks.

    \subsection{E3: Prithvi embeddings vs. hand-crafted features}

Table \ref{tab:Experiments_scores} (bottom), compares Prithvi foundation model embeddings against hand-crafted features using the optimal temporal configuration (S2 curr\_day + S1\_DESC\_closest). Prithvi embeddings combined with S1 achieved R²=0.514, matching the performance of traditional S2 indices + S1 (R²=0.514 from E1). Adding hand-crafted indices to Prithvi embeddings yielded marginal improvement (R²=0.515), representing only 0.2\% gain over the traditional approach. The Prithvi only (+ ERA5) configuration (without S1 or optical indices) underperformed the baseline S2\_curr\_day + S1\_DESC result, confirming that learned embeddings alone cannot replace the complementary information provided by SAR backscatter.

The negligible improvement from Prithvi embeddings suggests foundation models offer limited added value for point-based soil moisture regression with sparse training data. Embeddings were extracted from 224×224 pixel patches and averaged spatially, potentially diluting the signal from the 16×16 center patch encompassing station locations. With only 113 stations, high-dimensional embeddings (768 features) may introduce overfitting relative to compact hand-crafted indices ($\sim$40 features) that encode domain-specific moisture relationships. Expanding to larger ISMN datasets or extracting center-pixel embeddings from higher-resolution patches may improve foundation model performance.

\begin{table}[hbt]
	\centering
	\caption{Performance metrics for different sensor combinations.}
	\label{tab:Experiments_scores}
	\begin{tabular}{l
	                S[table-format=1.3]
	                S[table-format=1.4]
	                S[table-format=1.4]}
		\toprule
		\multicolumn{1}{c}{\textbf{Dataset}} &
		\multicolumn{1}{c}{\textbf{$R^{2}$}} &
		\multicolumn{1}{c}{\textbf{RMSE}} &
		\multicolumn{1}{c}{\textbf{MAE}} \\

        \cmidrule(lr){1-1} \cmidrule(lr){2-2}\cmidrule(lr){3-3} \cmidrule(lr){4-4}
		% \midrule

        S2\_curr\_day                     &   0.5066   &    0.0788    &    0.0606    \\
    
        S1\_ASC\_curr\_day                &   0.4037    &   0.0923     &    0.071    \\
        S1\_DESC\_curr\_day               &   0.4338    &  0.0943      &   0.0716     \\
        S1\_BOTH\_curr\_day               &   0.4057    &   0.0954     &    0.0724    \\
    
        S2\_closest + S1\_ASC\_closest    & 0.4809 & 0.0857 & 0.0633     \\
        S2\_closest + S1\_DESC\_closest   & 0.4869 & 0.0822 & 0.0633     \\
        S2\_closest + S1\_BOTH\_closest   & 0.4775 & 0.0823 &  0.0636   \\

        S1\_ASC\_curr\_day + S2\_closest  &   0.4616    &  0.0826      &    0.0637    \\
        S1\_DESC\_curr\_day + S2\_closest &   0.5098   &   0.0822     &   0.0629   \\
        S1\_BOTH\_curr\_day + S2\_closest &   0.4827   &    0.0837    &    0.0639    \\
         
        S2\_curr\_day + S1\_ASC\_closest  &   0.5009    &  0.0787      &  0.0604     \\
        S2\_curr\_day + S1\_DESC\_closest &   0.5142   &   0.0785    &  \textbf{0.0599}       \\
        S2\_curr\_day + S1\_BOTH\_closest &   0.5062    &   0.0787   &   0.0605  \\

        % \midrule

        % S2\_curr\_day + S1\_DESC + ERA5\_10day & 0.5185 & 0.0790 & 0.0590 \\

        \midrule
        Prithvi\_S2 & 0.5091 & 0.0789 & 0.0613\\
        Prithvi\_S2 + S1\_DESC\_closest &   0.5144   &   0.0785    &  0.0608       \\
        Prithvi\_S2 + indices + S1\_DESC\_closest &   \textbf{0.5151}   &  \textbf{0.0784} &  0.0607       \\

        \bottomrule

	\end{tabular}
\end{table}

\begin{figure}[hbt]
   \centering
   \includegraphics[width=\linewidth]{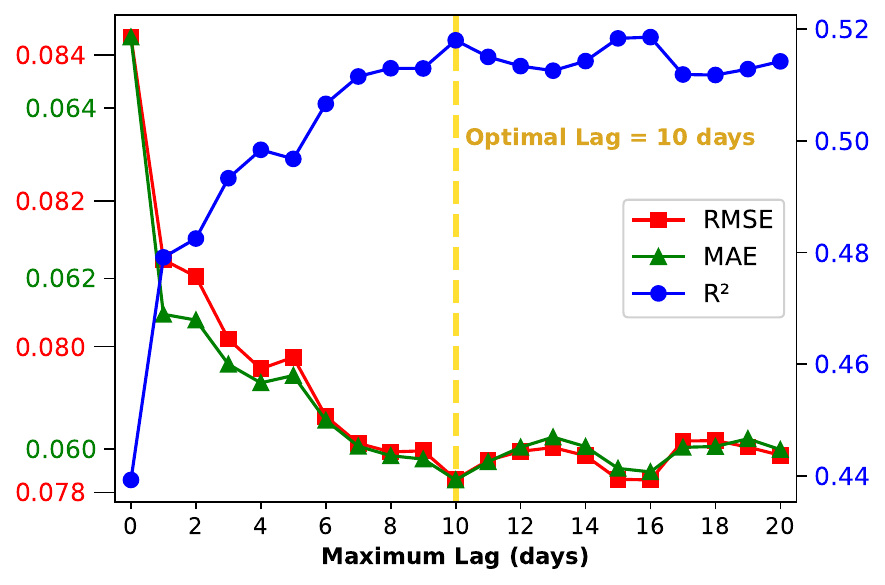}
   \caption{Model performance as a function of ERA5 temporal lag.}\label{fig:ERA5_lags_opt}
\end{figure}

\section{Conclusion}

This study evaluated multimodal machine learning approaches for high-resolution soil moisture estimation using 113 ISMN stations across Europe. Three systematic experiments identified optimal configurations for operational deployment. E1 demonstrated that hybrid temporal matching strategies—S2 current-day acquisitions with S1 descending closest—outperform uniform temporal approaches, with descending orbit SAR consistently superior to ascending. E2 revealed that 10-day ERA5 lookback window captures soil moisture dynamics most effectively, outperforming both shorter and longer integration periods across the 0-20 day range tested.
% E2 revealed that 10-day ERA5 lookback window captures both recent precipitation and antecedent moisture conditions more effectively than shorter 7-day or longer 20-day periods, exceeding typical surface soil memory timescales. 
E3 showed that Prithvi foundation model embeddings provide negligible improvement (R²=0.515) over traditional hand-crafted features (R²=0.514), indicating that domain-specific indices like NDVI and NDWI effectively encode moisture-relevant information at comparable computational efficiency.

\section{Acknowledgements}

This research is funded by the European Union’s Horizon Europe Programme under Grant Agreement No. 101135422 (UNIVERSWATER)

\small
\bibliographystyle{IEEEtranN}
\bibliography{references}

\end{document}